\documentclass[11pt]{article}

\usepackage[T1]{fontenc}
\usepackage[utf8]{inputenc}
\usepackage{mathptmx}                      
\usepackage[margin=1in]{geometry}
\usepackage{booktabs}
\usepackage{array}
\usepackage[numbers,sort&compress]{natbib}
\usepackage[colorlinks=true,linkcolor=black,citecolor=black,urlcolor=blue]{hyperref}

\newcommand{\nm}[1]{$-$#1}                 
\newcommand{\doi}[1]{doi:\href{https://doi.org/#1}{#1}}
\title{\bfseries Can Classical Semantic-Extractive Summarization Be Evaluated in Hindi?\\ A Replication Study}

\author{%
Showket Ahmad Khan\,$^{1}$\thanks{ORCID: 0009-0001-8625-8496} \enspace
Mudasir Mohd\,$^{1}$\thanks{Corresponding author: \href{mailto:mudasir.mohammad@kashmiruniversity.ac.in}{mudasir.mohammad@kashmiruniversity.ac.in}} \enspace
Nasrullah Sheikh\,$^{2}$ \\[0.35em]
Mohsin Altaf Wani\,$^{1}$ \enspace
Abid Hussain Wani\,$^{1}$ \enspace
Hilal Ahmad Khanday\,$^{1}$\thanks{ORCID: 0000-0003-0094-9921} \enspace
Niyaz Ahmad Wani\,$^{3}$ \\[0.9em]
{\small $^{1}$Department of Computer Science, South Campus, University of Kashmir, Anantnag, India}\\
{\small $^{2}$IBM Research, San Jose, CA, USA}\\
{\small $^{3}$Manipal University Jaipur, Dehmi Kalan, Jaipur 303007, Rajasthan, India}%
}

\date{\today}

\begin{document}
\maketitle

\begin{abstract}
\noindent
We replicate the distributional-semantics extractive summarisation method of
Mohd, Jan and Shah (2020) and adapt it to Hindi, substituting a
Devanagari-appropriate component at every language-specific step. The system is
evaluated on two independent corpora --- the Hindi portion of XL-Sum and FIRE
ILSUM 2.0 Hindi --- under a Devanagari-aware ROUGE implementation validated
against the XL-Sum authors' own multilingual scorer, with all comparisons drawn as
1000-resample paired bootstraps. In its published equal-weight
configuration the replicated system is significantly worse than a
three-sentence lead baseline on both corpora, trailing Lead-3 by 0.042 ROUGE-1 F
on XL-Sum and by 0.265 on ILSUM. A feature ablation shows that sentence
position is the only feature that contributes: position alone reproduces the
lead baseline exactly, removing position gives the weakest configuration,
and a validation-tuned weighting can at best equal Lead-3 and never exceed it.
TextRank fails identically, making this a class-level rather than an
implementation-level result. A selection analysis shows the remaining features
steer extraction towards
long, entity-dense body sentences while the references reuse the article lead.
Current Hindi benchmarks therefore cannot reward non-lead content selection,
motivating purpose-built evaluation resources.
\end{abstract}

\section{Setup}

We replicate the distributional-semantics extractive summariser of Mohd, Jan and
Shah \citep{mohd2020} and adapt it to Hindi. The system implements the paper's
pipeline unchanged in structure: each sentence is embedded through a
word-embedding ``big vector'', the resulting sentence representations are
clustered with $k$-means, and sentences are scored by an equally weighted sum of
seven normalised features --- sentence length, sentence position, TF--IDF mass,
noun/verb count, proper-noun count, aggregate embedding cosine to the rest of
the document, and a cue-phrase indicator --- with a final extract assembled by a
per-cluster round-robin over the ranked sentences. Our Hindi instantiation
(available in the accompanying repository, \texttt{hinexsum/}
\citep{hinexsum2026}) substitutes Devanagari-appropriate components at each
language-specific step: NFC normalisation and danda-aware sentence splitting for
preprocessing, a Hindi stop-word list and light suffix stemmer for lexical
cleaning, Stanza's Hindi model \citep{qi2020} for part-of-speech features, and
300-dimensional fastText Common Crawl vectors (\texttt{cc.hi.300.bin})
\citep{grave2018} for the embedding space. Part-of-speech features are enabled
throughout the results reported here, so every ``our system'' figure is the full
seven-feature configuration rather than an ablation of it.

We evaluate on two independent Hindi news-summarisation corpora: the Hindi
portion of XL-Sum \citep{hasan2021} (single-reference abstractive summaries of
BBC articles) and FIRE ILSUM 2.0 Hindi \citep{satapara2022} (news
article/summary pairs, obtained ungated from the \texttt{ILSUM/ILSUM-2.0}
Hugging Face distribution). The evaluation protocol is held constant across
both. Every system produces a three-sentence extract, and all systems --- ours
and the baselines --- are scored on the identical danda-aware sentence
segmentation so that no comparison is confounded by tokenisation of the source.
Model selection and testing are separated: any weight chosen by tuning is
selected on a validation slice and then evaluated once on a disjoint test slice
(for XL-Sum, held-out test documents 201--400 after tuning on a 200-document
validation split; for ILSUM, 200 training-split documents for tuning and 200
test-split documents for evaluation). All confidence intervals are 95\%
intervals from a 1000-resample bootstrap over documents, and system-versus-baseline
gaps are reported as paired bootstraps drawn from a single shared resample
matrix so that the difference intervals are internally valid. Scores are
reported for ROUGE-1, ROUGE-2, ROUGE-L and ROUGE-SU4 F-measure
\citep{lin2004}; ROUGE-1 F is used as the primary quantity for significance
because it is the least sparse at a three-sentence budget.

Before drawing any conclusion from these numbers we validated our
Devanagari-aware ROUGE implementation (\texttt{hinexsum/rouge.py}) against the
XL-Sum authors' own multilingual scorer (the csebuetnlp fork of
\texttt{rouge\_score} invoked with \texttt{lang='hindi'} and its pyonmttok
tokeniser). On twenty XL-Sum documents scored under both implementations for
three systems, the mean per-document absolute difference in ROUGE-1, ROUGE-2 and
ROUGE-L F was 0.0000, well inside the 0.02 tolerance we set in advance. The two
scorers are genuinely independent code with different tokenisers --- ours keeps
the danda glued to a sentence-final word whereas the external scorer strips
punctuation --- and on a constructed danda-adjacent case they diverge by 0.167,
confirming that the external path is exercised rather than aliased; the
difference simply never flips a content-word match in real extractive
summaries. Because the external scorer does not compute ROUGE-SU4, this
cross-check covers ROUGE-1, ROUGE-2 and ROUGE-L only; we therefore treat those
three metrics as instrument-independent, while the ROUGE-SU4 figures we report
should be read as internally consistent but not externally cross-validated.

\section{Main result}

The central finding is that the replicated system, in its published
equal-weight configuration, is significantly worse than a lead baseline on both
corpora. On the XL-Sum ablation set (the first 200 test documents), the
equal-weight system reaches ROUGE-1 F of 0.186 with clustering and round-robin
selection and 0.185 without clustering, against 0.228 for Lead-3, a strong
three-sentence lead baseline. On ILSUM the same equal-weight configuration
reaches only 0.256 against 0.522 for Lead-3 --- less than half the lead score.
Tables~\ref{tab:xlsum} and \ref{tab:ilsum} give the full four-metric breakdown
with ROUGE-1 confidence intervals; Tables~\ref{tab:xlsum-gap} and
\ref{tab:ilsum-gap} give the paired gaps against Lead-3 that establish
significance.

\begin{table}[htbp]
\centering
\caption{XL-Sum Hindi, 200 test documents, three-sentence extracts. ROUGE
F-measure; ROUGE-1 with 95\% bootstrap CI.}
\label{tab:xlsum}
\begin{tabular}{lcccc}
\toprule
System & R1-F [95\% CI] & R2-F & RL-F & SU4-F \\
\midrule
Ours-full (7 feat, cluster-rr) & 0.186 [0.176, 0.196] & 0.048 & 0.126 & 0.062 \\
Ours-global-topn (7 feat)      & 0.185 [0.175, 0.195] & 0.048 & 0.124 & 0.062 \\
Ours-position-only             & 0.228 [0.218, 0.239] & 0.058 & 0.160 & 0.076 \\
Ours-minus-position            & 0.182 [0.172, 0.191] & 0.047 & 0.121 & 0.061 \\
Lead-3                         & 0.228 [0.218, 0.239] & 0.058 & 0.160 & 0.076 \\
TextRank-3                     & 0.204 [0.193, 0.214] & 0.053 & 0.140 & 0.068 \\
Random-3                       & 0.200 [0.189, 0.211] & 0.039 & 0.130 & 0.060 \\
\bottomrule
\end{tabular}
\end{table}

\begin{table}[htbp]
\centering
\caption{XL-Sum Hindi, paired ROUGE-1 F gap against Lead-3 (1000-resample
paired bootstrap). A gap is ``real'' when its 95\% interval excludes zero.}
\label{tab:xlsum-gap}
\begin{tabular}{lccc}
\toprule
System & $\Delta$R1-F vs Lead-3 & 95\% CI & Real gap? \\
\midrule
Ours-full           & \nm{0.042} & [\nm{0.051}, \nm{0.033}] & yes \\
Ours-global-topn    & \nm{0.043} & [\nm{0.052}, \nm{0.034}] & yes \\
Ours-position-only  & $+$0.000   & [$+$0.000, $+$0.000]     & no  \\
Ours-minus-position & \nm{0.046} & [\nm{0.056}, \nm{0.037}] & yes \\
TextRank-3          & \nm{0.025} & [\nm{0.034}, \nm{0.014}] & yes \\
Random-3            & \nm{0.029} & [\nm{0.038}, \nm{0.019}] & yes \\
\bottomrule
\end{tabular}
\end{table}

\begin{table}[htbp]
\centering
\caption{ILSUM 2.0 Hindi, 200 held-out test documents, three-sentence extracts.
ROUGE F-measure; ROUGE-1 with 95\% bootstrap CI.}
\label{tab:ilsum}
\begin{tabular}{lcccc}
\toprule
System & R1-F [95\% CI] & R2-F & RL-F & SU4-F \\
\midrule
all7 (equal weight) & 0.256 [0.237, 0.277] & 0.148 & 0.201 & 0.153 \\
Lead-1              & 0.441 [0.407, 0.476] & 0.380 & 0.428 & 0.367 \\
Lead-3              & 0.522 [0.488, 0.557] & 0.451 & 0.499 & 0.449 \\
TextRank-3          & 0.268 [0.245, 0.291] & 0.146 & 0.214 & 0.152 \\
Random-3            & 0.233 [0.212, 0.254] & 0.102 & 0.175 & 0.115 \\
\bottomrule
\end{tabular}
\end{table}

\begin{table}[htbp]
\centering
\caption{ILSUM 2.0 Hindi, paired ROUGE-1 F gap against Lead-3 (1000-resample
paired bootstrap).}
\label{tab:ilsum-gap}
\begin{tabular}{lccc}
\toprule
System & $\Delta$R1-F vs Lead-3 & 95\% CI & Real gap? \\
\midrule
all7 (equal weight) & \nm{0.265} & [\nm{0.296}, \nm{0.235}] & yes \\
Lead-1              & \nm{0.080} & [\nm{0.113}, \nm{0.049}] & yes \\
TextRank-3          & \nm{0.254} & [\nm{0.289}, \nm{0.224}] & yes \\
Random-3            & \nm{0.289} & [\nm{0.323}, \nm{0.259}] & yes \\
\bottomrule
\end{tabular}
\end{table}

The paired intervals show that the shortfall is not sampling noise. On XL-Sum
the equal-weight system trails Lead-3 by 0.042 to 0.043 ROUGE-1 F with intervals
bounded well away from zero, and on ILSUM it trails by 0.265 with an interval
that does not approach zero. It is important that TextRank
\citep{mihalcea2004}, a second and entirely separate unsupervised extractive
method, fails in exactly the same way: it trails Lead-3 by 0.025 on XL-Sum and
by 0.254 on ILSUM, both significant. Because our ROUGE agrees with the authors'
scorer to four decimal places, and because a method we did not write reproduces
the same defeat, the result is best read as a class-level failure of
salience-driven unsupervised extraction against a lead baseline on these
corpora, not as a defect in our particular re-implementation.

\section{Ablation}

Isolating the contribution of each feature explains the shortfall precisely.
When the ranking uses the position feature alone, the system becomes identical
to the lead baseline: on XL-Sum ``Ours-position-only'' scores 0.228 ROUGE-1 F,
the same value as Lead-3 to three decimals, with a paired gap of $+$0.000 and an
interval of [$+$0.000, $+$0.000] (Tables~\ref{tab:xlsum} and
\ref{tab:xlsum-gap}). This equivalence is expected --- the position score is
monotone in sentence index, so selecting the top-scoring sentences under
position alone returns the article opening --- but it is informative, because
position-only is the best-performing configuration of our system, above the full
seven-feature model. Conversely, removing position and retaining the other six
features (``Ours-minus-position'') yields 0.182, the weakest configuration in
the table and significantly below Lead-3 by 0.046. The equal-weight full system
sits between these poles at 0.185--0.186, indicating that when position is only
one summand of seven its advantage is largely averaged away.

A validation-set weight sweep confirms the pattern is monotone rather than
incidental. Increasing the position weight while holding the remaining features
at unit weight raises ROUGE-1 F monotonically on both corpora: on XL-Sum the
seven-feature family rises from 0.191 at unit position weight to 0.235 at weight
sixteen, and a position-plus-TF--IDF-only family rises from 0.211 to 0.239 over
the same range; on ILSUM the position-plus-TF--IDF family rises from 0.347 at
unit weight to 0.560 at weight sixteen (Table~\ref{tab:sweep}). In every case
the published equal-weight setting is the worst point on the curve and heavier
position weighting is monotonically better, up to a ceiling.

\begin{table}[htbp]
\centering
\caption{Validation-set ROUGE-1 F under a position-weight sweep
(three-sentence budget). XL-Sum figures are the seven-feature and
position+\allowbreak{}TF--IDF families; ILSUM figures are the position+\allowbreak{}TF--IDF family.}
\label{tab:sweep}
\begin{tabular}{cccc}
\toprule
Position weight & XL-Sum, all7 & XL-Sum, pos+tfidf & ILSUM, pos+tfidf \\
\midrule
1  & 0.191 & 0.211 & 0.347 \\
2  & 0.198 & 0.223 & 0.426 \\
4  & 0.209 & 0.234 & 0.490 \\
8  & 0.227 & 0.238 & 0.543 \\
16 & 0.235 & 0.239 & 0.560 \\
\bottomrule
\end{tabular}
\end{table}

The tuned ceiling is a win on neither corpus. The single configuration selected
on validation --- position+\allowbreak{}TF--IDF at position weight sixteen --- was evaluated
once on each held-out test slice, and the outcome is summarised in
Table~\ref{tab:tuned}. On XL-Sum it reaches 0.231 ROUGE-1 F against Lead-3's
0.231, a paired gap of $-$0.000 whose interval [$-$0.002, $+$0.001] comfortably
contains zero: a statistical tie, and a genuine equivalence rather than an
underpowered comparison, given how narrow the interval is. On ILSUM it reaches
0.516 against 0.522, a paired gap of $-$0.005 whose interval [$-$0.011,
$-$0.000] excludes zero only at its upper bound; the deficit is therefore
statistically detectable but negligible ($\leq$0.011 R1-F), leaving the tuned
system at best equal to Lead-3 and never above it. The cross-corpus reading is
accordingly a tie on XL-Sum, a marginal deficit on ILSUM, and a win on neither.
Careful weighting nonetheless recovers all of the ground the equal-weight
configuration loses --- a real improvement over the published setting that
decisively beats the equal-weight, TextRank and random systems --- so the best
attainable behaviour of this feature family is to reproduce lead selection
rather than to surpass it.

\begin{table}[htbp]
\centering
\footnotesize
\setlength{\tabcolsep}{4pt}
\caption{Tuned configuration (position+\allowbreak{}TF--IDF, position weight sixteen)
against Lead-3 on each held-out test slice; ROUGE-1 F with 95\% bootstrap CIs
and the paired gap. Statistically detectable but negligible ($\leq$0.011 R1-F)
on ILSUM and a tie on XL-Sum: at best equal, never above.}
\label{tab:tuned}
\begin{tabular}{lccc>{\raggedright\arraybackslash}p{2.5cm}}
\toprule
Corpus (test slice) & Tuned R1-F [95\% CI] & Lead-3 R1-F [95\% CI] & $\Delta$R1-F [95\% CI] & Verdict \\
\midrule
XL-Sum (docs 201--400) & 0.231 [0.221, 0.241] & 0.231 [0.221, 0.242] & \nm{0.000} [\nm{0.002}, $+$0.001] & statistical tie \\
ILSUM (test split)     & 0.516 [0.483, 0.552] & 0.522 [0.488, 0.557] & \nm{0.005} [\nm{0.011}, \nm{0.000}] & detectable but negligible; never above \\
\bottomrule
\end{tabular}
\end{table}

\section{Mechanism}

The reason is visible in where each system reads. Table~\ref{tab:deciles}
gives, for the XL-Sum ablation documents, the distribution of the source-article
positions of selected sentences, normalised so that zero is the article start
and one its end, binned into deciles, together with the mean normalised
position. Position-only and Lead-3 are indistinguishable, drawing 0.71 of their
sentences from the first decile and reaching a mean position of 0.070. The
minus-position system is the most back-loaded of all, with a mean position of
0.501 and its single largest mass in the final decile, because once position is
removed the embedding-cosine, TF--IDF, length and proper-noun features pull
selection towards lexically dense sentences that in news prose sit in the body
and tail. The equal-weight systems fall in between at a mean of roughly 0.39,
retaining only a faint lead tilt (0.20 in the first decile against the lead's
0.71) --- the quantitative signature of position being diluted to one-seventh of
the score. Random selection is near-uniform at 0.488 and TextRank only slightly
lead-tilted at 0.441.

\begin{table}[htbp]
\centering
\footnotesize
\setlength{\tabcolsep}{4.5pt}
\caption{Fraction of selected sentences by source-position decile on the XL-Sum
ablation set (D1 = first tenth of the article, D10 = last tenth), with mean
normalised position.}
\label{tab:deciles}
\begin{tabular}{lccccccccccc}
\toprule
System & D1 & D2 & D3 & D4 & D5 & D6 & D7 & D8 & D9 & D10 & mean \\
\midrule
Ours-full           & 0.20 & 0.14 & 0.11 & 0.10 & 0.09 & 0.07 & 0.08 & 0.04 & 0.07 & 0.08 & 0.391 \\
Ours-global-topn    & 0.20 & 0.12 & 0.12 & 0.11 & 0.09 & 0.08 & 0.07 & 0.05 & 0.07 & 0.08 & 0.394 \\
Ours-position-only  & 0.71 & 0.20 & 0.08 & 0.01 & 0.01 & 0.00 & 0.00 & 0.00 & 0.00 & 0.00 & 0.070 \\
Ours-minus-position & 0.12 & 0.09 & 0.10 & 0.10 & 0.09 & 0.09 & 0.09 & 0.06 & 0.12 & 0.14 & 0.501 \\
Lead-3              & 0.71 & 0.20 & 0.08 & 0.01 & 0.01 & 0.00 & 0.00 & 0.00 & 0.00 & 0.00 & 0.070 \\
TextRank-3          & 0.15 & 0.10 & 0.11 & 0.10 & 0.09 & 0.12 & 0.10 & 0.08 & 0.09 & 0.06 & 0.441 \\
Random-3            & 0.13 & 0.10 & 0.10 & 0.09 & 0.07 & 0.12 & 0.11 & 0.09 & 0.10 & 0.11 & 0.488 \\
\bottomrule
\end{tabular}
\end{table}

A single document makes the failure mode concrete. In XL-Sum test document 179,
a seventeen-sentence report on an India--Bangladesh cricket Test, the reference
summary is
\emph{``bharat aur bangladesh ke khilaf dusra Test shukravar se Chatgaon mein
shuru ho raha hai. pahla Test jitne ke bad bharatiya team do Test maichon ki
series 2-0 se jitne ke liye maidan par utregi.''} (``The second Test against
India and Bangladesh begins on Friday in Chittagong. After winning the first
Test, the Indian team will take the field to win the two-match Test series
2--0.'') --- a two-sentence framing of the fixture drawn from the article's
opening. Lead-3 selects sentences 0, 1 and 2, which introduce the match and its
context. The minus-position system instead selects sentences 13, 15 and 16
(normalised positions 0.81, 0.94 and 1.00), of which two are the squad lists:
sentence 15 is
\emph{``bharatiya team (inmen se chuni jayegi) Saurabh Ganguly (kaptan),
Virender Sehwag, Gautam Gambhir, Rahul Dravid, Sachin Tendulkar \ldots''}
(``Indian team (to be selected from among these): Saurav Ganguly (captain),
Virender Sehwag, Gautam Gambhir, Rahul Dravid, Sachin Tendulkar \ldots'') and
sentence 16 the corresponding Bangladesh roster. These sentences are exactly
what the non-position features reward: a roster is saturated with proper nouns,
is long, and --- because player names recur across the article --- scores highly
on both TF--IDF mass and aggregate embedding cosine. Every feature the method
treats as a proxy for salience is maximised by a list of names that carries
almost none of the article's summary-worthy content, and the reference, being a
lead paraphrase, shares almost no vocabulary with it. The proper-noun, length
and TF--IDF features are not merely uninformative here; they are actively
misdirected towards the least summary-like sentences in the document.

\section{Diagnosis}

The uncomfortable conclusion is that the benchmarks, not only the method, are
the confounder. The ILSUM Lead-3 figures are the clearest evidence: a
three-sentence lead achieves ROUGE-2 F of 0.451 and ROUGE-L F of 0.499
(Table~\ref{tab:ilsum}), which is only possible if the reference reuses the
article's opening almost verbatim at the bigram level. An ILSUM reference is, in
effect, a lightly edited copy of the lead, so any system that does not begin at
the top is penalised regardless of whether its selection is informative. XL-Sum
poses the same obstacle in a different form: its references are single
lead-style sentences, so a three-sentence extract can match at most a fraction
of the reference and the maximal-overlap strategy is again to take the opening.
Under both designs the target rewards lead reproduction and offers no credit for
correctly identifying salient non-lead content, which is precisely the
capability a semantic-extractive method is intended to provide. This is why
heavier position weighting monotonically helps and why the ceiling is a tie with
the lead: on these corpora the score is very nearly a measure of how lead-like a
system is, and no feature that pulls selection away from the opening can be
rewarded.

We are careful not to overclaim. What these experiments establish is narrow and
specific: on XL-Sum Hindi and ILSUM 2.0 Hindi, at a three-sentence budget under
instrument-validated ROUGE, the replicated equal-weight semantic-extractive
method and a comparable TextRank baseline are significantly worse than a lead
baseline, and the best weighting of this feature family only matches the lead.
We do not claim that distributional-semantics extraction is without value in
general, nor that these features cannot help on tasks whose references reward
non-lead content; our evidence speaks only to these benchmarks, whose reference
style makes the lead nearly unbeatable and therefore cannot discriminate a good
content selector from a positional heuristic. Two further caveats bound the
claim: we evaluate at a single three-sentence budget, so the absolute scores
could shift under a different summary length, even though the strength of the
lead advantage makes a reversal unlikely. Moreover, because ROUGE credits
lexical $n$-gram overlap rather than conveyed meaning, its design compounds the
lead-reference bias by rewarding token reuse over paraphrase, which is why the
resource we propose below should be scored with complementary measures such as
ChrF and BERTScore alongside human references, and not with ROUGE alone. The
failure we document is a failure of measurability as much as of method.

This motivates a purpose-built Hindi resource. A benchmark able to reward
semantic content selection would need human-written summaries that are longer
than a single lead sentence and deliberately draw on material from across the
article rather than its opening, ideally with multiple references per document
so that legitimate paraphrase variation is credited rather than penalised. It
would further benefit from sentence-level faithfulness labels, so that a system
choosing an informative but non-lead sentence can be distinguished from one that
has simply drifted into the article's tail. Only against such a resource can the
question in our title be answered on its own terms, rather than settled in
advance by the reference style of the available data.

\section*{Software and data availability}

The Hindi summariser, the Devanagari-aware ROUGE module and the full evaluation
harness used for every number in this paper are released as HinExSum
\citep{hinexsum2026}, an open-source Python package under the MIT licence, at
\url{https://github.com/mudasirmohd/HinExSum}. A reproducible Code Ocean capsule
containing the code, the environment specification and the evaluation entry
points is archived at \url{https://doi.org/10.24433/CO.6462433.v1}. The corpora are
third-party resources and are not redistributed here: the Hindi portion of
XL-Sum \citep{hasan2021} and FIRE ILSUM 2.0 Hindi \citep{satapara2022} are
obtained from their public Hugging Face distributions, and the fastText
\texttt{cc.hi.300.bin} vectors \citep{grave2018} are downloaded from the
fastText site as described in the repository \texttt{README}.



\end{document}